\documentclass[conference]{IEEEtran}
\IEEEoverridecommandlockouts

\usepackage{cite}
\usepackage{amsmath,amssymb,amsfonts}
\usepackage{algorithmic}
\usepackage{graphicx}
\usepackage{textcomp}
\usepackage{xcolor}
\usepackage{booktabs}
\usepackage{colortbl}
\usepackage{multirow}
\usepackage{makecell}
\usepackage{array}
\def\BibTeX{{\rm B\kern-.05em{\sc i\kern-.025em b}\kern-.08em
    T\kern-.1667em\lower.7ex\hbox{E}\kern-.125emX}}
\begin{document}

\title{NeuroLex: A Lightweight Domain Language Model for EEG Report Understanding and Generation\\

\thanks{This work was supported by the Institute for Information \& Communications Technology Planning \& Evaluation (IITP) grant funded by the Korea government (MSIT) (No. RS-2019-II190079, Artificial Intelligence Graduate School Program, Korea University).}
}

\author{\IEEEauthorblockN{~~~~~~~Kang Yin}
\IEEEauthorblockA{\textit{~~~~~~~Dept. of Artificial Intelligence} \\
\textit{~~~~~~~Korea University}\\
~~~~~~~Seoul, Republic of Korea \\
~~~~~~~charles\_kang@korea.ac.kr}

\and
\IEEEauthorblockN{~~Hye-Bin Shin}
\IEEEauthorblockA{\textit{~~~~Dept. of Brain and Cognitive Engineering} \\
\textit{Korea University}\\
Seoul, Republic of Korea \\
hb\_shin@korea.ac.kr}


}

\maketitle


\begin{abstract}
Clinical electroencephalogram (EEG) reports encode domain-specific linguistic conventions that general-purpose language models (LMs) fail to capture.  
We introduce \textbf{NeuroLex}, a lightweight domain-adaptive language model trained purely on EEG report text from the Harvard Electroencephalography Database.  
Unlike existing biomedical LMs, NeuroLex is tailored to the linguistic and diagnostic characteristics of EEG reporting, enabling it to serve as both an independent textual model and a decoder backbone for multimodal EEG–language systems.  
Using span-corruption pretraining and instruction-style fine-tuning on report polishing, paragraph summarization, and terminology question answering, NeuroLex learns the syntax and reasoning patterns characteristic of EEG interpretation.  
Comprehensive evaluations show that it achieves lower perplexity, higher extraction and summarization accuracy, better label efficiency, and improved robustness to negation and factual hallucination compared with general models of the same scale.  
With an EEG-aware linguistic backbone, NeuroLex bridges biomedical text modeling and brain–computer interface applications, offering a foundation for interpretable and language-driven neural decoding.

\end{abstract}

\begin{IEEEkeywords}
brain--computer interface, electroencephalogram, language model;
\end{IEEEkeywords}

\section{INTRODUCTION}
Electroencephalogram (EEG) has long served as a non-invasive window into brain activity, powering a wide spectrum of brain–computer interface (BCI) applications—from disease detection~\cite{b3, lee2020_pilotcnn} and cognitive workload~\cite{intro:cognitive-workload} monitoring to emotion recognition~\cite{b1,ding2013_smoking, intro:emotion} and intention decoding~\cite{b2,lee2015_motion}. Yet despite significant advances in EEG decoding algorithms, a persistent gap remains in how such neural data are described, summarized, and communicated in natural language. EEG reports, routinely written by neurologists, encapsulate domain knowledge that bridges low-level signal phenomena (e.g., rhythmic slowing, spikes, asymmetries) with high-level clinical interpretations. They represent not merely annotations but \textit{structured linguistic codifications of human neurophysiological reasoning}.

However, most current BCI research largely overlooks this linguistic layer. While deep neural networks~\cite{dl, lee2003_svm, bulthoff2003_bmcv} have achieved notable performance in EEG classification and cross-modal alignment~\cite{intro:chatgptbci, b4,lee1996_wavelet}, these systems are typically paired with generic large language models (LLMs) such as GPT or T5 that lack grounding in EEG-specific semantics~\cite{intro:emotion-copilot, suk2011_freqbands}. Consequently, the generated textual outputs—whether diagnostic summaries, data explanations, or multimodal captions—often sound fluent but lack clinical precision. For example, expressions like \textit{“attenuation of alpha rhythm”} follows domain conventions that general models rarely internalize. This mismatch limits interpretability, interoperability, and the practical deployment of BCI systems in clinical or research settings.

In recent years, there has been a growing demand for domain-specialized yet lightweight language models that can operate efficiently in laboratory pipelines~\cite{biobert,lee1995_multilayer,lee1999_cascade} and embedded neurotechnology environments~\cite{bio-pretraining}. Unlike massive general LLMs, compact EEG-domain models could be directly integrated into closed-loop systems, on-device analysis, or multimodal training frameworks without prohibitive computational cost~\cite{b5,lee2018_curling}. From a scientific standpoint, such models are also crucial for EEG–language alignment: enabling textual supervision for EEG representation learning, automatic report generation for neurodiagnostic datasets, and precise language grounding for multimodal BCI systems.

To address these gaps, we propose \textbf{NeuroLex}, a lightweight domain-adaptive language model pretrained purely on EEG-related texts. NeuroLex is designed as both a standalone linguistic backbone and a decoder foundation for multimodal EEG–text integration. Built upon the encoder–decoder transformer framework of T5~\cite{t5}, it undergoes (1) \textbf{domain-adaptive pretraining (DAPT)} using span corruption on large-scale de-identified EEG report corpora, and (2) \textbf{supervised fine-tuning (SFT)} on task-specific objectives such as summarization, polishing, and terminology question answering. Through these two complementary stages, NeuroLex learns not only the vocabulary and syntax of EEG reporting but also the structured reasoning embedded in real clinical narratives.

Besides, this study empirically tests four hypotheses on EEG-domain language modeling for BCI research:
\begin{itemize}
    \item H1 (Domain Adaptation): DAPT yields lower perplexity and higher terminology coverage than general T5.
    \item H2 (Task Benefit): DAPT + SFT models outperform general or single-stage models on EEG extraction and summarization tasks.
    \item H3 (Data Efficiency): domain adaptation improves learning stability and efficiency under limited labeled data.
    \item H4 (Robustness): DAPT models generalize better, reducing terminology hallucinations and negation errors.
\end{itemize}

By providing an EEG-aware linguistic backbone, NeuroLex bridges the gap between biomedical text modeling and practical BCI needs, forming a foundation for interpretable and linguistically grounded brain–language systems.

\section{METHODS}
\begin{figure}[t]
    \centering
    \resizebox{0.9\linewidth}{!}{
        \includegraphics{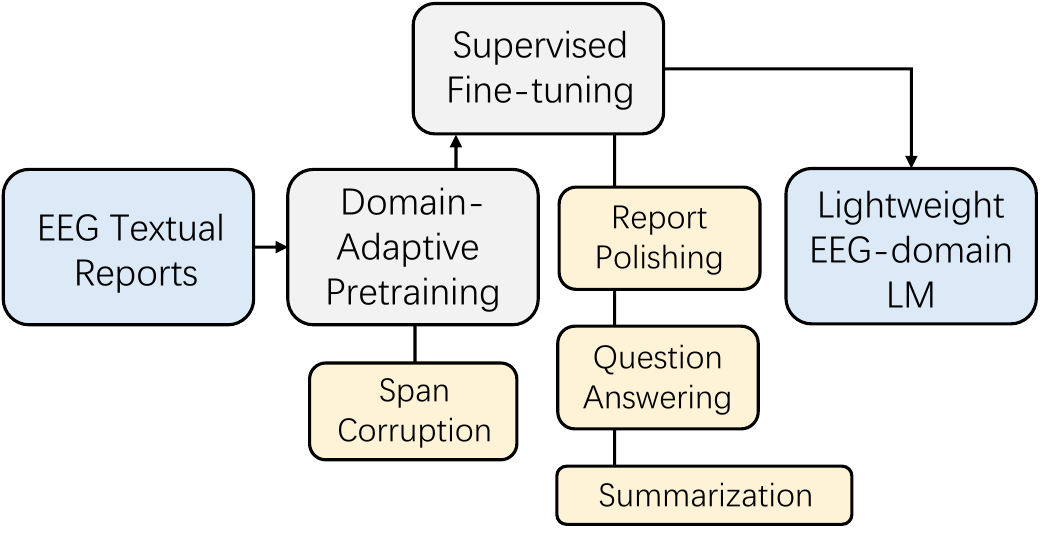}
    }
    \caption{Overview of the NeuroLex training pipeline with a two-stage training process: DAPT to learn EEG-specific linguistic structure, and SFT to adapt the model toward practical text understanding and generation tasks.}
    \label{framework}
\end{figure}

\subsection{Overview}
NeuroLex follows the encoder–decoder transformer architecture of T5-Base and is trained exclusively on textual data. As shown in Fig.~\ref{framework}, the training process comprises two stages: DAPT to learn EEG-specific linguistic structure, and  SFT to adapt the model toward practical text understanding and generation tasks, as suggested by S. Gururangan \textit{et al.}~\cite{dapt_sft}.

\subsection{Domain-Adaptive Pretraining}
\subsubsection{Objective}
we employ the span-corruption objective, but differently we mask \textit{only EEG-specific terminology}, where $\sim$15~\% of tokens in each input sequence are replaced with sentinel tokens (e.g., $<$extra\_id\_0$>$) and the model learns to reconstruct the missing spans.
This approach encourages the model to capture long-range context, structured co-occurrence, and reporting conventions typical of clinical EEG language.

\subsubsection{Corpus}
all pretraining data are derived exclusively from the Harvard Electroencephalography Database (HEEDB)~\cite{dataset}, a large-scale open-access repository of clinical EEG recordings accompanied by textual reports.
\begin{table}[t]
\centering
\caption{Illustrative examples of the four training objectives.}
\setlength{\tabcolsep}{4pt}
\begin{tabular}{@{}p{\columnwidth}@{}}
\toprule
\textbf{Task: Span Corruption} \\
\textbf{Input:} The background shows \texttt{\textless extra\_id\_0\textgreater} with diffuse slowing.\\
\textbf{Output:} \texttt{\textless extra\_id\_0\textgreater} posterior-dominant alpha rhythm.\\
\midrule
\textbf{Task: Report Polishing} \\
\textbf{Input:} Mild slow waves seen bilateral temporal region.\\
\textbf{Output:} Mild slowing is observed over the bilateral temporal regions.\\
\midrule
\textbf{Task: Question Answering} \\
\textbf{Input:} What does FIRDA indicate in an EEG?\\
\textbf{Output:} Frontal intermittent rhythmic delta activity, often associated with diffuse cerebral dysfunction.\\
\midrule
\textbf{Task: Summarization} \\
\textbf{Input:} Frequent spike-and-wave discharges over the left temporal area, with occasional right temporal involvement. Background rhythm remains reactive.\\
\textbf{Output:} Recurrent left temporal epileptiform discharges with preserved reactivity.\\
\bottomrule
\end{tabular}
\vspace{-1mm}
\label{tab:examples}
\end{table}
We extract the full set of EEG reports and perform extensive cleaning and normalization, including:
(a) removal of protected health information and non-EEG sections, and (b) standardization of spacing, punctuation, and segmentation.

The resulting corpus contains approximately 70K EEG reports and around 800K paragraphs, covering both findings and impression styles but unified as continuous text for pretraining.
This ensures that the model learns general EEG report phrasing and terminology such as spike-and-wave discharges, focal slowing, alpha attenuation, and triphasic waves.

\subsection{Supervised Fine-Tuning}
The fine-tuning stage adapts NeuroLex to instruction-style text generation tasks.
We design three complementary supervised objectives, each formulated as text-to-text mapping, using LLM-generated pseudo-labels to create training supervision from the same corpus.

\subsubsection{Report Polishing}
the model is trained to rewrite unstructured or noisy EEG sentences into fluent, standardized forms while preserving clinical meaning.
This task teaches grammatical refinement and style normalization.
\subsubsection{Question Answering (QA)}
we automatically generate short QA pairs covering EEG terminology, phenomena, and interpretive phrases. This task enhances domain reasoning and factual recall.
\subsubsection{Summarization}
each EEG paragraph is summarized into a concise one-to-two-sentence description.
LLM-generated summaries serve as pseudo ground truths, guiding the model to compress technical EEG text while retaining key findings.

Table~\ref{tab:examples} presents illustrative examples for the four training objectives.  
All objectives are formulated under the unified T5 text-to-text paradigm and optimized with standard cross-entropy loss.  
Training inputs are truncated to a maximum length of 512 tokens, and outputs to 256 tokens.  
We start with pretrained Flan-T5 Base, and both pretraining and fine-tuning are performed on three NVIDIA A6000 GPUs with a batch size of 32, a learning rate of $2\times10^{-4}$, and up to five epochs using early stopping based on validation perplexity.

\section{EXPERIMENTS}

We evaluate NeuroLex from both intrinsic and extrinsic perspectives to examine how domain adaptation and fine-tuning influence its linguistic and clinical performance. 
All experiments use reports from HEEDB with an 80/10/10 train/validation/test split. 
In addition, we manually curate approximately 1K test samples for each task, where LLM-generated references are verified by human annotators to ensure factual correctness. 
Unless otherwise noted, all reported results are computed on this manually curated test set. All models share the same parameter scale as Flan-T5 Base for fair comparison. 
Our implementation is publicly available at https://github.com/Kang1121/NeuroLex.
\subsection{Intrinsic Evaluation}
\begin{table}[t]
\vspace{-1mm}
\centering
\renewcommand{\arraystretch}{0.9} 
\caption{Intrinsic evaluation on tokenizer quality using different training corpora.}
\begin{tabular}{lrrrr}
\toprule
\textbf{Tokenizer} & \textbf{OOV (\%)} & \textbf{AS} & \textbf{SS (\%)} & \textbf{MTR} \\
\midrule
Flan-T5 Base & 49.04 & 3.38 & 80.23 & 1.75\\
EEG-based & 10.19 & 1.96 & 40.70 & 1.06  \\
\bottomrule
\end{tabular}
\vspace{-1mm}
\label{tab:tokenizer}
\vspace{-1mm}
\end{table}

\begin{table}[t]
\vspace{-1mm}
\centering
\renewcommand{\arraystretch}{0.9} 
\setlength{\tabcolsep}{1.8mm}
\caption{Intrinsic evaluation of domain-adaptive pretraining. Lower PPL indicates better language modeling.}
\begin{tabular}{lrrrr}
\toprule
\textbf{Model} & \textbf{PPL (All)}  & \textbf{PPL (Imp.)} & \textbf{Top-1 (\%)}  &  \textbf{Top-5 (\%)} \\
\midrule
Flan-T5 Base & 16.98 & 5.92 &2.60& 2.90\\
+\,DAPT & 805.08 & 562.23 &72.10& 82.30\\
+\,SFT  & \textbf{5.73} & \textbf{4.68} &56.00& 64.50\\
\rowcolor{gray!30}
+\,DAPT\,→\,SFT & 6.08 & 5.29 &\textbf{74.60}& \textbf{84.20} \\
\bottomrule
\vspace{-1mm}
\end{tabular}
\label{tab:intrinsic}
\vspace{-1mm}
\end{table}



We first evaluate whether domain-adaptive pretraining improves the lexical representation of EEG-report language.
As shown in Table~\ref{tab:tokenizer}, the EEG-domain tokenizer greatly enhances coverage and consistency, reducing the out-of-vocabulary (OOV) rate from 49.04 \% to 10.19 \%, and cutting both the average subwords per token (AS) and single-word split ratio (SS) by nearly half.
The multi-word token ratio (MTR) approaches 1.0, showing that EEG-specific expressions are now preserved as single, coherent units.

Next, we assess intrinsic linguistic competence via perplexity (PPL) and masked-span reconstruction accuracy.
Table~\ref{tab:intrinsic} shows that SFT alone achieves the lowest PPL on both the full corpus (5.73) and Impression (Imp.) sections (4.68), reflecting better fluency, while the DAPT→SFT model attains the highest reconstruction accuracy (Top-1 = 74.60 \%, Top-5 = 84.20 \%), demonstrating stronger contextual understanding and domain sensitivity.


\subsection{Information Extraction}
Having established its linguistic grounding, we next evaluate whether NeuroLex can transform that understanding into structured reasoning.  
EEG report structuring is framed as a text-to-text information extraction (IE) task: each sentence is converted into a JSON-like output with five attributes—\textit{Laterality}~(Lat.), \textit{Localization}~(Loc.), \textit{Pattern}~(Patt.), \textit{Frequency}~(Freq.), and \textit{Negation}~(Neg.).  
A hybrid rule–dictionary tagger provides weak labels, while a small human-verified subset (1K samples) serves as the test set.  

As summarized in Table~\ref{tab:ie}, DAPT$\rightarrow$SFT achieves the highest overall F1 (0.575), outperforming both single-stage variants.  
Improvements are most pronounced in \textit{Localization} (0.621) and \textit{Frequency} (0.678), dimensions that require contextual interpretation rather than direct lexical matching.  
The results confirm that domain adaptation enhances factual precision and consistency in structured EEG information extraction.

\begin{table}[t]
\vspace{-1mm}
\centering
\renewcommand{\arraystretch}{0.9} 
\caption{Slot-level F1 scores for EEG information extraction.}
\begin{tabular}{lcccccc}
\toprule
\textbf{Model} & \textbf{Lat.} & \textbf{Loc.} & \textbf{Patt.} & \textbf{Freq.} & \textbf{Neg.} & \textbf{Avg.} \\
\midrule
Flan-T5 Base & 0.197 & 0.147 & 0.122 & 0.225 & 0.442 & 0.227  \\
+\,DAPT & \textbf{0.615} & 0.505 & 0.239 & 0.225 & 0.442 & 0.405  \\
+\,SFT  & 0.527 & 0.543 & 0.280 & 0.517 & 0.551 & 0.484  \\
\rowcolor{gray!30}
+\,DAPT\,→\,SFT & 0.484 & \textbf{0.621} & \textbf{0.401} & \textbf{0.678} & \textbf{0.693} & \textbf{0.575} \\
\bottomrule
\vspace{-1mm}
\end{tabular}
\label{tab:ie}
\vspace{-1mm}
\end{table}

\begin{table}[t]
\vspace{-1mm}
\centering
\renewcommand{\arraystretch}{0.9} 
\caption{Extrinsic evaluation on summarization consistency.}
\begin{tabular}{lccc}
\toprule
\textbf{Model} & \textbf{ROUGE-L} & \textbf{BERTScore} & \textbf{Fact-F1} \\
\midrule
Flan-T5 Base & 0.214 & 0.848 & 0.742   \\
+\,DAPT & 0.103 & 0.807 & 0.736  \\
+\,SFT  & 0.695 & 0.955 & \textbf{0.942} \\
\rowcolor{gray!30}
+\,DAPT\,→\,SFT & \textbf{0.707} & \textbf{0.956} & 0.941 \\
\bottomrule
\vspace{-1mm}
\end{tabular}
\vspace{-1mm}
\label{tab:summarization}
\end{table}

\subsection{Summarization}
We further test whether NeuroLex can condense long EEG paragraphs into concise, faithful statements~\cite{factual}.  
Unlike conventional findings to impression mappings, each paragraph is summarized independently.  Evaluation metrics include ROUGE-L for lexical overlap, BERTScore~\cite{bertscore} for semantic similarity, and a rule-based Fact-F1 that penalizes semantic contradictions.

Table~\ref{tab:summarization} shows that the DAPT$\rightarrow$SFT configuration produces the most balanced results (ROUGE-L = 0.707, BERTScore = 0.956 and Fact-F1 = 0.941).  
Qualitatively, its summaries preserve negations and avoid hallucinated terms, indicating that domain pretraining improves both linguistic fluency and factual safety in clinical summarization.

\subsection{Data Efficiency}

\begin{figure}[t]
\vspace{-1mm}
\centering
\includegraphics[width=0.71\columnwidth]{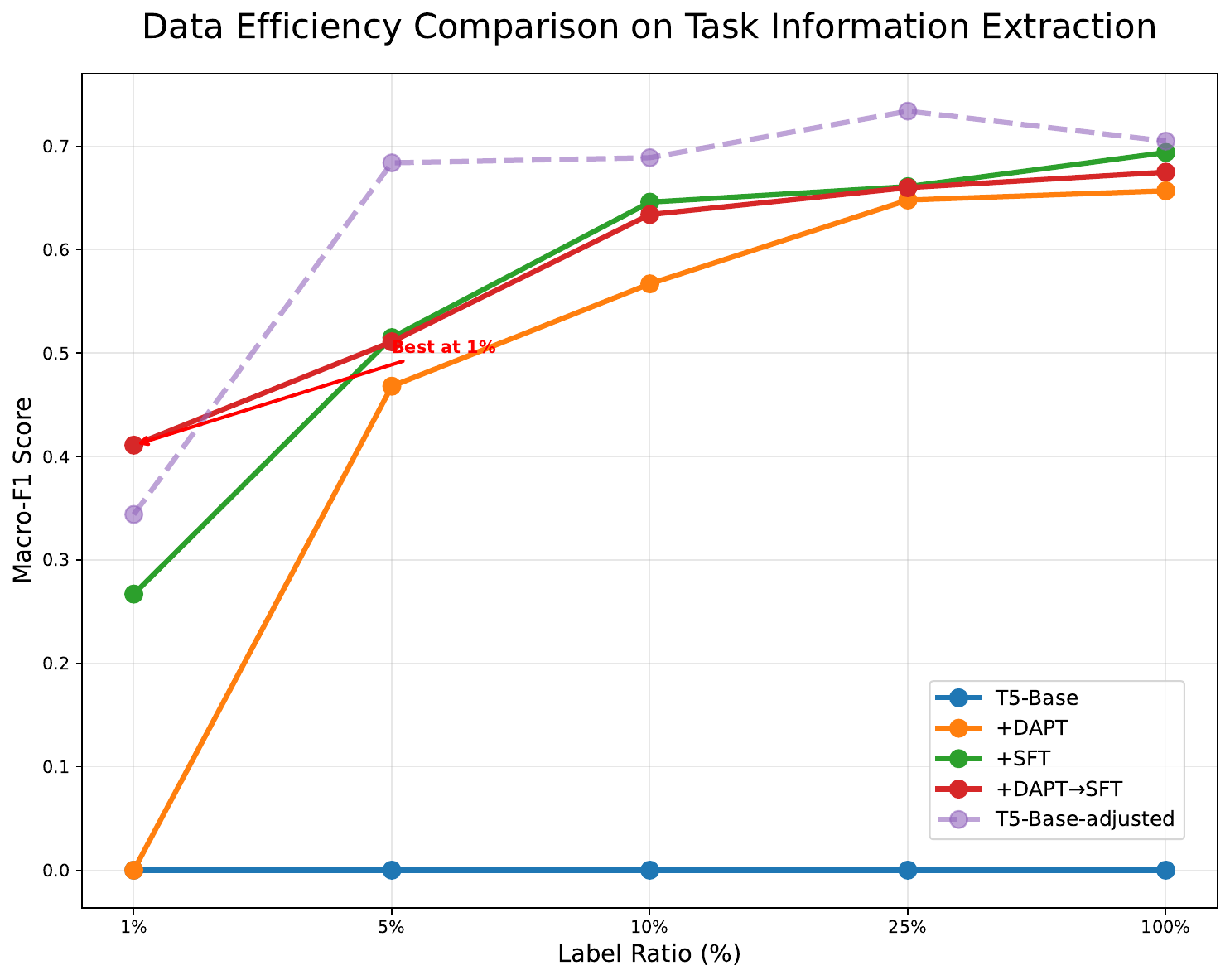} 
\caption{A comparison of traditional feature-based EEG report generation and our proposed end-to-end description generation with model understanding.}
\label{fig:dataeff}
\vspace{-1mm}
\end{figure}

Because EEG annotation is expensive, we examine how each model scales with limited supervision using the same information-extraction task.  
Training data are subsampled to 1~\%, 5~\%, 10~\%, 25~\% of available labels, and Macro-F1 scores are reported in Table~\ref{tab:dataeff} with learning curves shown in Fig.~\ref{fig:dataeff}.  
Even with only 1~\% of labeled data, the DAPT$\rightarrow$SFT variant achieves 0.411 Macro-F1, outperforming all baselines and reaching saturation earlier.  
This indicates that exposure to large amounts of unlabeled EEG text provides strong inductive bias, greatly improving data efficiency.

\begin{table}[t]
\centering
\renewcommand{\arraystretch}{0.9} 
\caption{Data-efficiency comparison on the IE task (Macro-F1).}
\begin{tabular}{lccccc}
\toprule
\textbf{Label Ratio} & \textbf{1~\%} & \textbf{5~\%} & \textbf{10~\%} & \textbf{25~\%} & \textbf{100~\%} \\
\midrule
Flan-T5 Base & 0.000 & 0.000 & 0.000 & 0.000 & 0.000 \\
+\,DAPT & 0.000 & 0.468 & 0.567 & 0.648 & 0.657 \\
+\,SFT  & 0.267 & \textbf{0.515} & \textbf{0.646} & \textbf{0.661} & \textbf{0.694} \\
\rowcolor{gray!30}
+\,DAPT\,→\,SFT & \textbf{0.411} & 0.511 & 0.634 & 0.660 & 0.675 \\
\bottomrule
\vspace{-1mm}
\end{tabular}
\vspace{-1mm}
\label{tab:dataeff}
\end{table}

\subsection{Robustness and Clinical Reliability}



We assess robustness under distribution shifts by perturbing negation cues (e.g., ``no'' $\rightarrow$ ``without''), scope, and double negatives, and evaluate with: negative-adversarial F1~(Neg-Adv F1, classification accuracy under negation), expected and maximum calibration error~(ECE/MCE, confidence calibration), term-precision~(Term-Prec., ratio of newly introduced EEG terms; lower is better), and contradiction rate~(Contr-Rate, fact-level conflicts; lower is better). 

As shown in Table~\ref{tab:robustness}, the DAPT$\rightarrow$SFT model attains the best overall balance, achieving the highest Neg-Adv F1 (0.683) and a low Contr-Rate (0.058) while maintaining reasonable calibration.  
It remains stable to negation variations and rarely introduces unseen terminology, indicating that domain adaptation improves both reliability and factual consistency.



\begin{table}[t]
\vspace{-1mm}
\centering
\setlength{\tabcolsep}{1.5mm}
\renewcommand{\arraystretch}{0.9} 
\caption{Robustness to negation, noise, and template shifts (lower is better except Neg-Adv F1).}
\begin{tabular}{lrrrrr}
\toprule
\textbf{Model} & \textbf{Neg-Adv F1} & \textbf{ECE} & \textbf{MCE} & \textbf{Term-Prec} & \textbf{Contr-Rate}\\
\midrule
Flan-T5 Base                  & 0.336 & 0.508 & 0.728 & \textbf{0.002} & 0.118  \\
+\,DAPT                  & 0.080 & \textbf{0.155} & \textbf{0.349} & 0.150 & 0.367  \\
+\,SFT                   & 0.668 & 0.683 & 0.830 & 0.103 & \textbf{0.052}  \\
\rowcolor{gray!30}
+\,DAPT\,→\,SFT & \textbf{0.683} & 0.671 & 0.809 & 0.106 &  0.058\\
\bottomrule
\vspace{-1mm}
\end{tabular}
\label{tab:robustness}
\vspace{-1mm}
\end{table}

\section{CONCLUSIONS}

We presented \textbf{NeuroLex}, a lightweight EEG-domain language model trained through domain-adaptive pretraining and supervised fine-tuning on EEG reports.  
Experiments show that NeuroLex achieves stronger linguistic understanding, higher task accuracy, better data efficiency, and greater robustness than general models of the same size.  
With an EEG-aware textual backbone, it lays the groundwork for interpretable and language-driven brain–computer interface research.



\end{document}